\documentclass[twocolumn]{article}

\usepackage{dcolumn}
\usepackage{graphicx}
\usepackage{subfigure}
\usepackage{bm}
\usepackage{multirow}
\usepackage{multicol}
\usepackage{lipsum}
\usepackage[fleqn]{amsmath}
\usepackage{abstract}
\usepackage{float}
\usepackage[T1]{fontenc}
\usepackage[utf8]{inputenc}
\usepackage{authblk}
\usepackage{cite}
\usepackage{authblk}

\title{\textbf{MulGAN: Facial Attribute Editing by Exemplar}}

\author[a]{\textbf{Jingtao Guo}}

\author[a]{\textbf{Zhenzhen Qian}}

\author[a]{\textbf{Zuowei Zhou}}

\author[a]{\textbf{Yi Liu}\thanks{Corresponding author. Tel.: +86 135-2082-4503. E-mail address: yiliu@bjtu.edu.cn.}}

\affil[a]{\rm{\textbf{School of Computer and Information Technology, Beijing Jiaotong University, Beijing, China}}}

\par  
\affil[a]{\rm{\{jingtaoguo, qianzz, zhouzw, yiliu\}@bjtu.edu.cn} }

\date{}
\begin{document}
		
	\twocolumn[
	\maketitle 
	\begin{onecolabstract}		
	Recent studies on face attribute editing by exemplars have achieved promising results due to the increasing power of deep convolutional networks and generative adversarial networks. These methods encode attribute-related information in images into the predefined region of the latent feature space by employing a pair of images with opposite attributes as input to train model, the face attribute transfer between the input image and the exemplar can be achieved by exchanging their attribute-related latent feature region. However, they suffer from three limitations: (1) the model must be trained using a pair of images with opposite attributes as input; (2) weak capability of editing multiple attributes by exemplars; (3) poor quality of generating image. Instead of imposing opposite-attribute constraints on the input image in order to make the attribute information of images be encoded in the predefined region of the latent feature space, in this work we directly apply the attribute labels constraint to the predefined region of the latent feature space. Meanwhile, an attribute classification loss is employed to make the model learn to extract the attribute-related information of images into the predefined latent feature region of the corresponding attribute, which enables our method to transfer multiple attributes of the exemplar simultaneously. Besides, a novel model structure is designed to enhance attribute transfer capabilities by exemplars while improve the quality of the generated image. Experiments demonstrate the effectiveness of our model on overcoming the above three limitations by comparing with other methods on the CelebA dataset.
	\end{onecolabstract}
	]

\section{Introduction}
\begin{figure*}[ht]
	\centering
	\includegraphics[width=16cm]{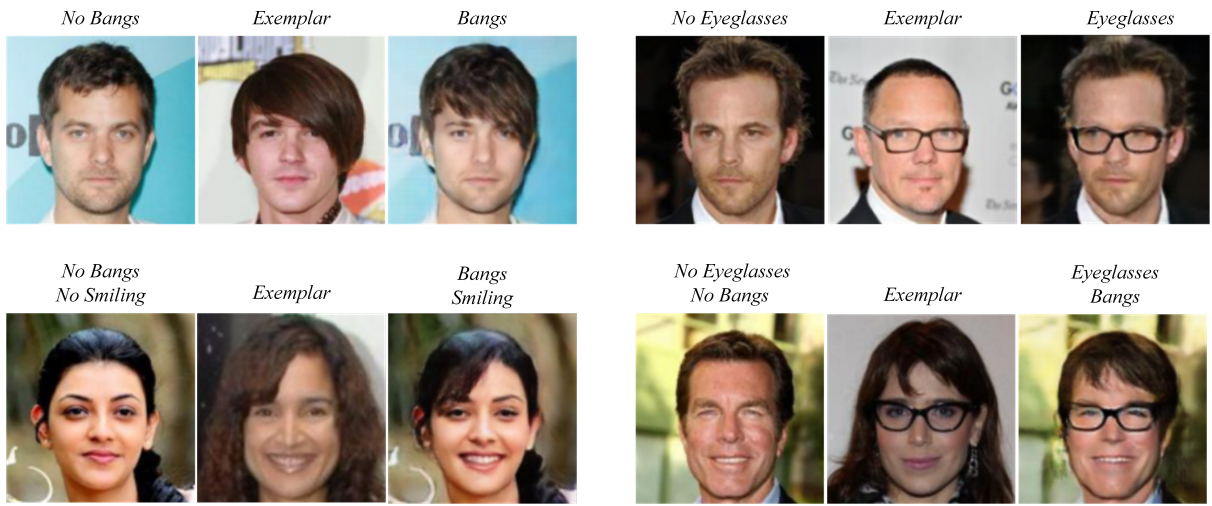}
	\caption{Our MulGAN generates the results of facial attribute editing by exemplar.~\label{F1}}
\end{figure*}

This work investigates the facial attribute editing by exemplar, which aims to transfer the single or multiple facial attributes of interest (e.g., smiling, bangs and male) in the exemplar to the source face image, while preserve as much attribute-excluding details as possible. Usage of exemplar images allows more precise specification of desired modifications and improves the diversity of conditional image generation. As shown in Fig.~\ref{F1}, the exactly the same style of attributes are transferred from the exemplar to the source image while keeping the attribute-excluding details unchanged.

Recently, many deep learning based methods~\cite{1,2,3,4,5,6,7,8,9,10} have emerged and achieved promising result for the facial attribute transferring task due to the rapid development of deep convolutional networks, especially generative adversarial networks~\cite{11}. These methods can be divided in two categories, the attribute labels based ones and the exemplar based ones. In this paper, we focus on the exemplar based methods~\cite{5,6,7}, which adopt the prevalent convolutional encoder-decoder network architecture and are trained with using pairs of images with opposite attributes to encode attribute-related information in images into the predefined region of the latent feature space. These methods can successfully make the attributes be encoded in a disentangled manner in the latent space. Especially for single facial attribute transferring task, these works are proven to effectively transfer a single facial attributes of interest from the exemplar to the source one.

Unfortunately, these methods suffer from the following limitations. First, these methods encode attribute-related information in images to the predefined region of the latent feature space by imposing opposite-attribute constraints on the input image. which causes only one facial attributes of the exemplar to be transmitted at a time. Very limited ability for generalization to transferring multiple attributes simultaneously during testing phase. Another alternative way for transferring multiple attributes is through multiple transferring operations where transferring only one attribute of exemplar at a time, which is not only time-consuming and cumbersome, but also the effect is not guaranteed too. Second, these methods adopt encoder-decoder structure, where more spatial pooling or down-sampling are used to obtain higher level abstract representation, which can enhance the model generation capability and attribute manipulation capability. However, higher level abstract representation means more fine details lost, loss of attribute-related fine details weakens capability of transferring attributes by exemplars, loss of attribute-independent details causes poor quality of new generating image.

\begin{figure*}[ht]
	\centering
	\includegraphics[width=16cm]{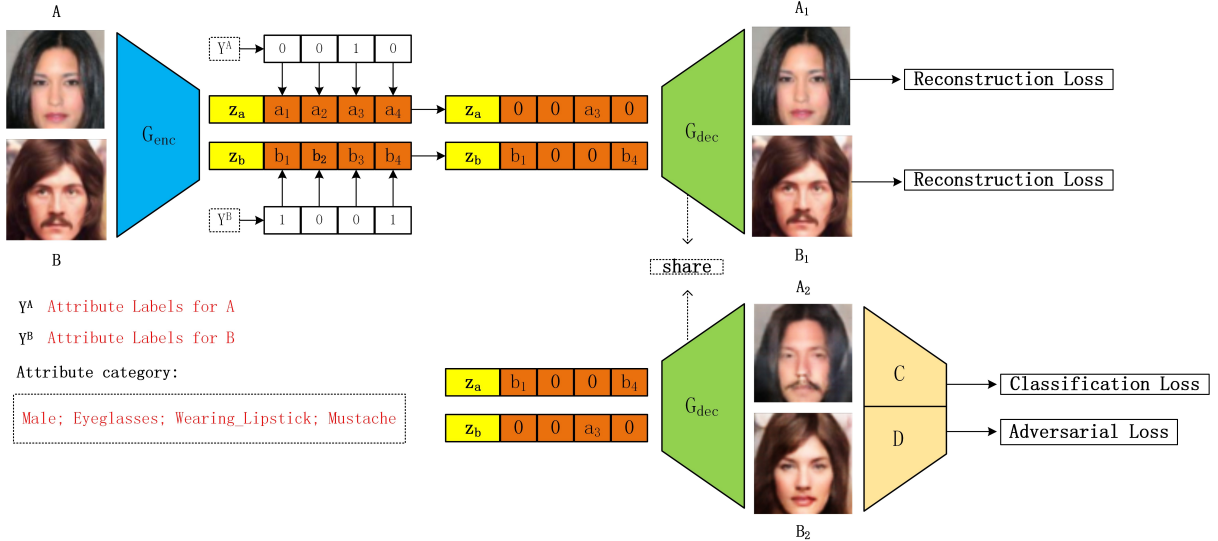}
	\caption{The MulGAN model architecture, which contains three main components: a generator (encoder-decoder architecture), a discriminator and an attribute classifier. The generator is used for the facial attribute editing task, the attribute classifier is used to extract the attribute-related information of images and the discriminator is employed for visually realistic generation.~\label{F2}}
\end{figure*}
To overcome these limitations, we present a novel convolutional encoder-decoder generative network to transfer multiple face attributes of the exemplar simultaneously. As shown in Fig.~\ref{F2}, our method splits the latent encodings of an image into two equal parts: attribute-relevant part and attribute-irrelevant part along the channel dimension. The attribute-irrelevant part preserve enough information for the recovery of the attribute-excluding details. The attribute-relevant part is used to accurately extract the style of attributes in the images, and the attribute-relevant part is again divided into different blocks, where each block encodes the information of a single attribute. Then, we directly apply the attribute labels constraint to the predefined attribute-related blocks in the latent feature space. As shown in Fig.~\ref{F2}, the predefined attribute-related blocks are filtered through binary attribute labels before being sent to the decoder, which makes the predefined attribute-related blocks own the original attribute label information. The predefined attribute-related blocks become a new, stronger representation capability, learnable “labels”. when we employ an attribute classifier to constrain the generated image to own the desired attributes, which forces the model learn to extract the attribute-related information of images into the predefined attribute-related blocks of each attribute. Finally, less down-sampling convolutional layers are used in our method to reduce the loss of fine details, which is important to enhance attribute manipulation ability by exemplars and improve image quality. More attribute-related details are retained to enhance attribute transfer capabilities by exemplars, more attribute-independent details are retained to improve the quality of the generated image. These components cooperate with each other not only enables our method to transfer multiple attributes of the exemplar simultaneously. but also improve the quality of the generated image and enhance attribute transfer capabilities by exemplars.

\section{Related Work}
\subsection{Generative Adversarial Networks}
Generative adversarial networks (GANs)~\cite{11} originally aims to generate images from random noise by adversarial training, which consists of a generator and a discriminator. The discriminator $D$ aims to learns to distinguish the generated images from the real ones, while the generator $G$ aims to try its best to generate convincing samples are indistinguishable from real ones. The adversarial process is formulated as a minimax game as
\begin{eqnarray}
\min_{\substack{G}} \max_{\substack{D}}& &\bm{E}_{x\sim p_{data}(x)}[\log(D(x))]+\nonumber\\
									   & &\bm{E}_{z\sim p_{z}(z)} [\log(1-D(G(z)))]
\end{eqnarray}
where the distribution of the real sample $x$ is denoted by $p_{data}(x)$, and $p_{z}(z)$ the distribution of the noise input $z$.

However, typical GAN suffers from adversarial training instability and mode collapse problem. Initially DCGAN~\cite{12} improved GAN to solve training instability through CNN and batch normalization~\cite{13}, however, the effect is limited. Recently, a number of methods~\cite{14,15,16,17} have been developed to improve the adversarial training stability and alleviate mode collapse problem. In this work, we employ WGAN~\cite{14} for the adversarial learning, WGAN uses the Wasserstein-1 distance for comparing the generated and real data distributions
\begin{eqnarray}
\min_{\substack{G}} \max_{\substack{\lVert D \rVert_{L\leq1}}} & &\bm{E}_{x\sim p_{data}(x)}[D(x)]-\nonumber\\
&&\bm{E}_{z\sim p_{z}(z)} [D(G(z))]
\end{eqnarray}
where $D$ is constrained to be the 1-Lipschitz function, which can be implemented by imposing a gradient penalty on the discriminator.

\subsection{Facial Attribute Editing}
Recently, many deep learning based facial attribute editing methods~\cite{1,2,3,4,5,6,7,8,9,10} have emerged following the introduction of generative adversarial networks and the encoder-decoder architecture. These methods can be divided in two categories, the attribute labels based ones and the exemplar based ones. The former edit the attributes of face image conditioned on the given attribute labels. Several attribute labels based methods have been proposed for multiple facial attribute editing. Guim et al.~\cite{18} proposed invertible Conditional GANs (icgan) by using encoders to inverse the mapping of a cGAN~\cite{19}. The attribute editing is achieved by encoding an image into the latent representation using the encoder, then decoding the representation conditioned on the desired attribute labels using the cGAN. StarGAN~\cite{8} achieves the facial attribute editing based on the attribute labels by introducing attribute classification loss~\cite{20} and cycle consistency loss, which can perform multiple attribute editing simultaneously using only a single model. Adopting a similar strategy, He et al.~\cite{9} proposed AttGAN that applies the attribute classification constraint to the generated image instead of the strict attribute-independent constraint on the latent representation, which guarantees the correct change of the attributes by decoding the representation conditioned on the desired attribute labels. In order to improve image quality and enhance attribute editing capacity. STGAN~\cite{10} further improves AttGAN by taking the difference attribute vector instead of target attribute vector as attribute labels, which pay more attention to edit the attributes to be changed. In addition, a novel selective transfer module is designed for adaptively select and modify latent feature layer by layer. However, the attribute labels based methods suffer from the following limitations: (1) Do not allow users to make more precise specification of desired modifications; (2) Editing results lacks of richness and diversity;(3) Poor editing performance for attributes with complex structures. This is mainly because the information provided by binary attribute labels is very limited, however, the style of attributes in images does vary. These methods use attribute labels directly for guiding face attribute editing, which fails to generate complex and diverse, more precise specification of attribute style.

To overcome these limitations of the attribute labels based methods, some the exemplar based methods have emerged and achieved promising results. GeneGAN~\cite{5} learns disentangled attribute representation from pairs of images with opposite attributes by adversarial training, which can perform fine-grained editing by transferring the style of attribute from the exemplar, like manipulating the "eyeglasses" attribute by swapping the "eyeglasses" latent representation with the exemplar, to generate novel images wearing eyeglasses with the same style of attributes in the exemplar. However, GeneGAN only can transfer a single attribute by exemplar. For different attribute, it has to trains different models, which is cumbersome and unfriendly for practical applications. DNA-GAN~\cite{6} can learn disentangled representation for multiple attributes simultaneously by only using a single model. The disentangled latent representation for each attribute is extracted by adversarial training with pairs of images with opposite attributes. At same the time, the iterative training strategy is adopted to achieve disentangle multiple attributes only using a single model. However, DNA-GAN suffers from the poor quality of generated image. In order to improve the quality of generated image, ELEGANT~\cite{3} further improves DNA-GAN by introducing multi-scale discriminators and adopting the residual learning strategy. However, ELEGANT improves image quality at the cost of weakened attribute transferring capabilities by exemplars. Although ELEGANT and DNA-GAN can manipulate several attributes simultaneously by exemplar in theory. In the experiment, we found that this ability is very limited, they can't balance well the quality of generated images with the ability to transfer attributes by exemplar.

In this paper, we proposed MulGAN, a novel, scalable method for single or multiple facial attribute editing by exemplar. which is mostly motivated by DNA-GAN, ELEGANT and StarGAN, we mainly focus on the disadvantages of these three methods on modeling the relation between the latent representation and the attributes, and propose a novel method to solve such problem.

\section{Method}
In this section, we propose the MulGAN approach for transferring multiple facial attributes from an exemplar. As shown in Fig.~\ref{F2}, it consists of three sub-networks: a generator (encoder-decoder architecture), a discriminator and an attribute classifier. The generator is used for the facial attribute editing task, the attribute classifier is used to extract the attribute-related information of images and the discriminator is employed to ensure the new generated image visual reality. Details of the design principles and the loss function are described below.

\subsection{Model}
Let $A$ be the input image with $n$ binary attributes $Y^{A} = [y_{1}^{A},\cdots, y_{n}^{A}]$ .$B$ is the exemplar image with the corresponding binary attributes $Y^{B} = [y_{1}^{B},\cdots, y_{n}^{B}]$,$B$ is generated by randomly shuffling $A$. The encoder $G_{enc}$ maps $A$ and $B$ into latent feature representation $Z^{A}$ and $Z^{B}$,respectively, $Z^{A} = G_{enc}(A)$, $Z^{B} = G_{enc}(B)$. Then, MulGAN first split the latent feature representation from the encoder into two equal parts (attribute-relevant and attribute-irrelevant) along the channel dimension. The attribute-irrelevant part preserves enough information for the recovery of the attribute-excluding details, the attribute-relevant part is used to accurately extract the style of attributes in the images. 
\begin{equation}
[a, z_{a}] = Z^{A}, [b, z_{b}] = Z^{B}
\end{equation}

\noindent
where $a$ is called the attribute-relevant part, and $z_{a}$ is called the attribute-irrelevant part. The same thing applies for $Z^{B}$.

And the attribute-relevant part is again divided into different blocks, where each block is used to encode a single attribute. As shown in Fig.~\ref{F2}.
\begin{equation}
[a_{1},\cdots,a_{i},\cdots,a_{n}] = a, [b_{1},\cdots,b_{i},\cdots,b_{n}] = b
\end{equation}

\noindent
where $a_{i}$ is supposed to map the $y_{i}^{A}$, the $i$-th attribute in the label. $n$ is the total number of the attributes. The same thing applies for $b$ .

Although we allocate a specific block for each transferred attribute, which cannot guarantee that each attribute is encoded into different blocks. The encoder has to be trained to make the specific attribute information in the image to encode into the corresponding attribute-relevant block. The previous methods learn such disentangled representations by employing pairs of images with opposite attribute, which trains the model with respect to a particular attribute each time. We break this restriction by employing the binary attribute labels and introducing an attribute classifier. As shown in Fig.~\ref{F2}, for the image $A$, the attribute-relevant blocks $[a_{1},\cdots,a_{i},\cdots,a_{n}] = a$ are filtered through binary attribute labels $Y^{A}$ before being sent to the decoder $G_{dec}$, when $y_{i}^{A}$ is equal to 0, the corresponding attribute-relevant block $a_{i}$ is set to 0, $y_{i}^{A}$ is equal to 1, the corresponding attribute encoding remains unchanged. After this operation, the $n$ different attribute-relevant blocks are in one-to-one correspondence with $n$ transferred attributes, which not only makes attribute-relevant blocks contain the original binary attribute label information, but also own a stronger representation ability (the binary attribute labels can only represent whether the facial image has a certain attribute, however, the attribute-relevant blocks can not only indicate whether the face image has a certain attribute, but also can represent the style of attributes in the face image). when we employ an attribute classifier $C$ to constrain the generated image to own the desired attributes, which forces the attribute-relevant blocks to extract the attribute-relevant information of image for each attribute. For example, the “eyeglasses” attribute is true for facial image, the attribute-relevant block for “eyeglasses” can encode the structural and texture information of the “eyeglasses” in the image. The same thing applies for attribute-relevant blocks $[b_{1},\cdots,b_{i},\cdots,b_{n}] = b$ .

\begin{equation}
a^{'} = [a_{1} \times y_{1}^{A}, \cdots, a_{i} \times y_{i}^{A}, \cdots, a_{n} \times y_{n}^{A}]
\end{equation}

\begin{equation}
b^{'} = [b_{1} \times y_{1}^{B}, \cdots, b_{i} \times y_{i}^{B}, \cdots, b_{n} \times y_{n}^{B}]
\end{equation}

\begin{equation}
Z^{A^{'}} = [a^{'}, z_{a}], Z^{B^{'}} = [b^{'}, z_{b}]
\end{equation}

Then we exchange the attribute-irrelevant part in latent encodings so as to obtain two new latent representations.

\begin{equation}
Z^{C} = [b^{'}, z_{a}], Z^{D} = [a^{'}, z_{b}]
\end{equation}

We expect that $Z^{C}$ is the encoding of the image $A$ version with exactly the same style of attributes in the $B$ image, and $Z^{D}$ the encodings of the image $B$ version with exactly the same style of attributes in the $A$ image.
Then via a decoder $G_{dec}$, we can get four newly generated images.

\begin{equation}
A_{1} = G_{enc}(Z^{A^{'}}), B_{1} = G_{enc}(Z^{B^{'}})
\end{equation}

\begin{equation}
A_{2} = G_{enc}(Z^{C}), B_{2} = G_{enc}(Z^{D})
\end{equation}

\noindent
where $A_{1}$ and $B_{1}$ are the direct reconstructions of $A$ and $B$, respectively, $A_{2}$ and $B_{2}$ are generated by swapping the attribute-relevant latent encodings.

\subsection{Loss Functions}

Face attribute editing by exemplar should only transfer the attribute-relevant information from the exemplar to the source image, while keeping the other details unchanged in the source. To this end, the reconstruction loss is introduced to make the attribute-irrelevant part of the latent feature space conserve enough information for the later recovery of the attribute-excluding details. The reconstruction losses between $A$ and $A_{1}$, ensure the quality of directly reconstructed samples.

\begin{equation}
L_{rec} = \lVert A - A_{1} \rVert
\end{equation}

\noindent
where the decoder $G_{dec}$ learn to best reconstruct the original image $A$ using the attribute-irrelevant part $z_{a}$ and the attribute-relevant blocks $a^{'}$ that are filtered through binary attribute labels $Y^{A}$.

The attribute classification loss. As mentioned above, we use the binary attribute labels to assist the encoder in encoding attribute-relevant blocks. The binary attribute labels clarify the one-to-one correspondence between n attributes and n the attribute-relevant blocks, which makes attribute-relevant blocks contain the original binary attribute label information. When we employ the attribute classification loss to constrain the generated image to own the desired attributes, which can forces the attribute-relevant blocks to extract the attribute-relevant information of image for each attribute.

For given the attribute-irrelevant part $z_{a}$ of source image $A$ and the attribute-relevant part $b^{'}$ of the exemplar $B$, we expect the generator to produce a new image $A_2$ with exactly the same style of attributes in the exemplar while keeping the attribute-excluding details the same as the source image. The attribute classification loss is uesd to optimize the generator $G$ on the generated $A_2$, formulated as follows,
\begin{eqnarray}
L_{cls_{g}} &=& \sum_{i=1}^{n}(-y_{i}^{A}\log C_{i}(A_{2}) -\nonumber\\
& & (1 - y_{i}^{A})\log (1-C_{i}(A_{2})))
\end{eqnarray}

\noindent
where, $C_{i}(A_{2})$ indicates the prediction of the $i$-th attribute for the generated images $A_2$. By minimizing this objective, the generator $G$ can encode each attribute into different blocks and generate the new image $A_{2}$ with the same style of attributes in the exemplar $B$.

The attribute classifier $C$ is trained on the input images $A$ with labeled attributes $Y^{A}$,  by the following objective,
\begin{eqnarray}
L_{cls_{c}} &=& \sum_{i=1}^{n}(-y_{i}^{A}\log C_{i}(A) - \nonumber\\
& & (1-y_{i}^{A})\log (1-C_{i}(A)))
\end{eqnarray}

\noindent
where $C_{i}(A)$ indicates the prediction of the $i$-th attribute for the input images $A$. By minimizing this objective, $C$ learns to classify a input image $A$ into its corresponding original attributes $Y^{A}$.

The adversarial learning between the generator (including the encoder and decoder) and discriminator is introduced to make the generated image $A_{2}$ visually realistic. Following WGAN~\cite{24}, the adversarial losses for the the discriminator and generator are formulated as below,
\begin{eqnarray}
L_{adv_d} &=& -E[D(A)]-E[D(B)]  \nonumber\\
&& +E[D(A_{2})]+ E[D(B_{2})]
\end{eqnarray}
\begin{eqnarray}
L_{adv_g} = -E[D(A_{2})] - E[D(B_2)]
\end{eqnarray}

Overall Objective. By combining the attribute classification constraint, the reconstruction loss and the adversarial loss, MulGAN can transfer multiple attributes simultaneously from the exemplar to the target image. Overall, the objective for the encoder and decoder is formulated as below,
\begin{equation}
L_G = L_{adv_g} + \lambda_{g}L_{cls_g} + \lambda_{rec}L_{rec}
\end{equation}

\noindent
and the objective for the discriminator and the attribute classifier is formulated as below
\begin{equation}
L_D = L_{adv_d}
\end{equation}
\begin{equation}
L_C = L_{cls_c}
\end{equation}

\noindent
Here, $\lambda_{rec}$ and $\lambda_{g}$ are weights to define the importance of different losses for the generator network.


\section{Implementation details}
\subsection{Network Architecture}
The encoder $G_{enc}$ consists of four down-sampling layers. and the decoder $G_{dec}$ consists of four up-sampling layers that recovers the image back to its original size. Unlike other methods that use more down-sampling convolutional layers to obtain higher level abstract representation. our network model only decreases the resolution four times, using strided convolutions to 1/8 of the original size, which is important to enhance attribute manipulation ability by exemplars (more attribute-related details are retained) and improve image quality. The discriminator $D$ has six down-sampling convolutional layers and a fully-connected layer followed by a sigmoid function to predict whether the image is real or fake. The classifier shares all convolutional layers with the discriminator, but owns a new fully-connected layer, followed by sigmoid functions to predict attributes. 

\subsection{Training Details}
MulGAN is implemented by the machine learning system Pytorch, and executed it on a computer with a single NVIDIA 1080Ti GPU (11GB), The network is trained using with a batch size of 32. The model is optimized using Adam~\cite{21} optimizer with a learning rate of 0.0001. The coefficients for the losses are set as: $\lambda_{g} = 10$ and $\lambda_{rec} = 100$, which balances the effects of different losses.


\section{Experiments}

\subsection{Benchmark Dataset}
We train the proposed model and evaluate its performance on the CelebA dataset~\cite{22}, which is consist of 200k facial images, each with 40 attributes annotation. We follow the standard protocol: splitting the dataset into 160k images for training, 20k for validation and 20k for testing. For the manipulation attributes by exemplar, we choose "Bangs”, “Eyeglasses”, “Gender”, “Smiling”, “Mustache”, “Blond Hair”, “Smiling”, “Pale Skin” and “Mouth Slightly Open”, which cover most attributes used in the existing works.

\begin{figure*}[htb]
	\centering
	\includegraphics[width=0.85\textwidth]{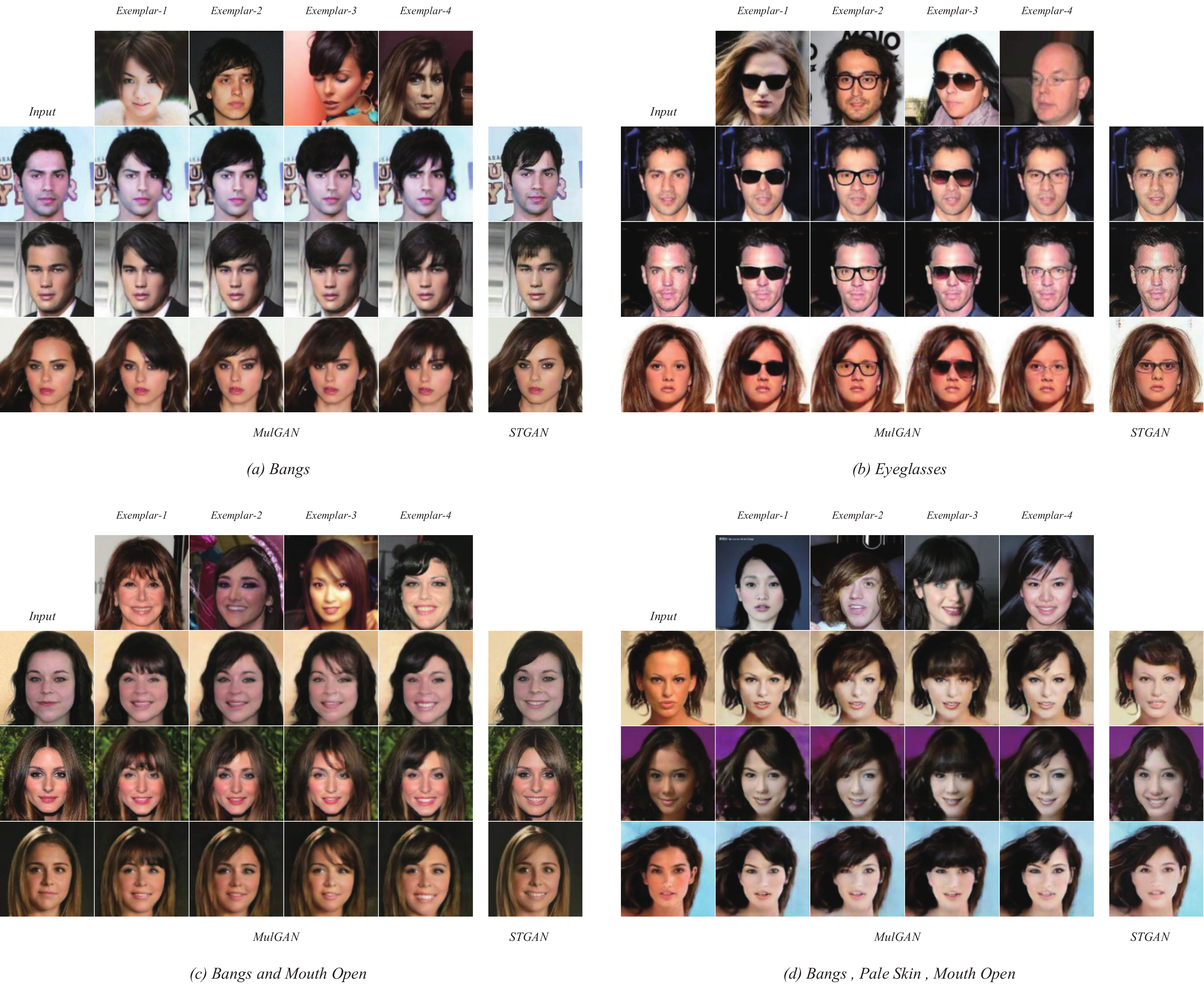}	
	\caption{Facial Attribute Editing by Exemplars. MulGAN can transfer exactly the same style of single or multiple facial attributes from the exemplar to the input image. Zoom in for better resolution.\label{F3}}
\end{figure*}
\subsection{Facial Attribute Editing by Exemplars}
In order to demonstrate that our model can generate face images by exemplars, we compare our results with STGAN, one of the best, representative and state-of-the-art approach based on attribute labels. MulGAN can generate different face images with exactly the same style of attribute in the exemplars, whereas STGAN is only able to generate a common style of attribute with the desired attribute labels. Instead of using binary attribute labels directly to guide image generation, MulGAN use the attribute labels to constraint to the predefined blocks of the latent feature space for each attribute, then the predefined attribute-related blocks become a new, stronger representation capability, learnable “label”, which can accurately extract the style of attribute information in the exemplar. As shown in Fig.~\ref{F3}.
\newcommand{\tabincell}[2]{\begin{tabular}{@{}#1@{}}#2\end{tabular}}

\begin{figure*}[htb]
	\centering
	\includegraphics[width=0.85\textwidth]{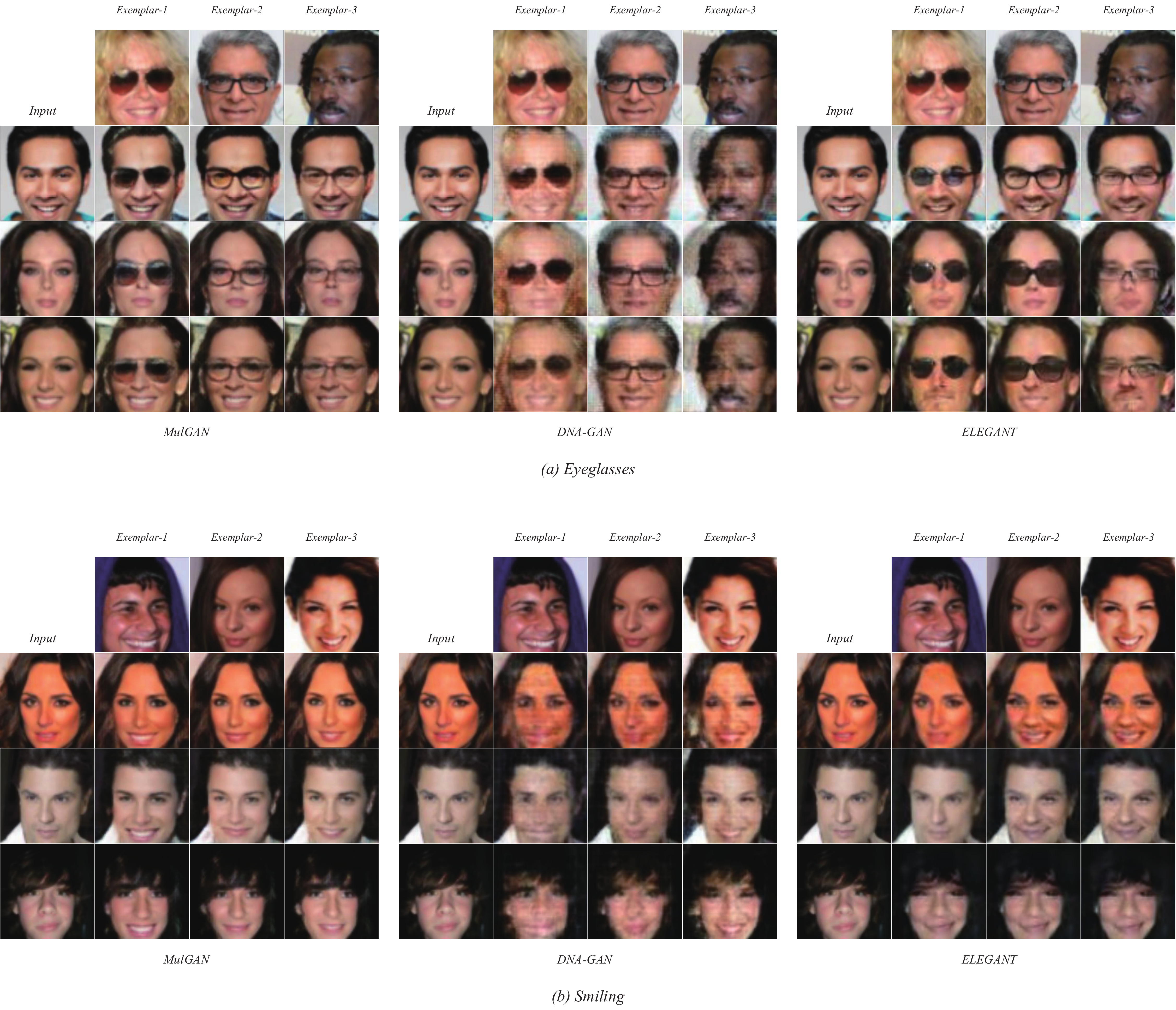}
	\caption{Comparisons of single facial attribute editing by exemplars. Panels from left to right are results of MulGAN, DNA-GAN and ELEGANT.\label{F4}}
\end{figure*}

\subsection{Comparison with Existing Work}
For single facial attribute editing by exemplars, we compare our MulGAN with DNA-GAN and ELEGANT. Three models are performed on the same face images and reference images with respect to two representative attributes (eyeglasses and smiling). As shown in Fig.~\ref{F4}, DNA-GAN has strong attribute migration capabilities by exemplars. However, DNA-GAN not only migrates the attribute-related information from the reference image to the target image, but also a large amount of attribute-irrelevant information is also transferred, which causes severe distortion of the generated image. ELEGANT is able to achieve better visual effects and reconstruction results due to the residual learning. However, ELEGANT improves image quality at the cost of weakened attribute migration capabilities by exemplars. As can be seen, in some cases ELEGANT fail to transfer the same style of attribute in the reference images to target image. Compared to DNA-GAN and ELEGANT, our MulGAN can accurately transfer the style of attribute in the reference images. credited to 1) the attribute labels constraint to the predefined attribute-related region in the latent feature space, which combine the attribute classification loss make the model can efficiently and accurately learn disentangled latent feature representations for all attributes; 2) MulGAN use less down-sampling convolutional layers to keep fine details of feature map, which is important to enhance attribute manipulation ability by exemplars and improve image quality. More attribute-related details are retained to enhance attribute transfer capabilities by exemplars, more attribute-independent details are retained to improve the quality of the generated image.

\begin{figure*}[htb]
	\centering
	\includegraphics[width=0.75\textwidth]{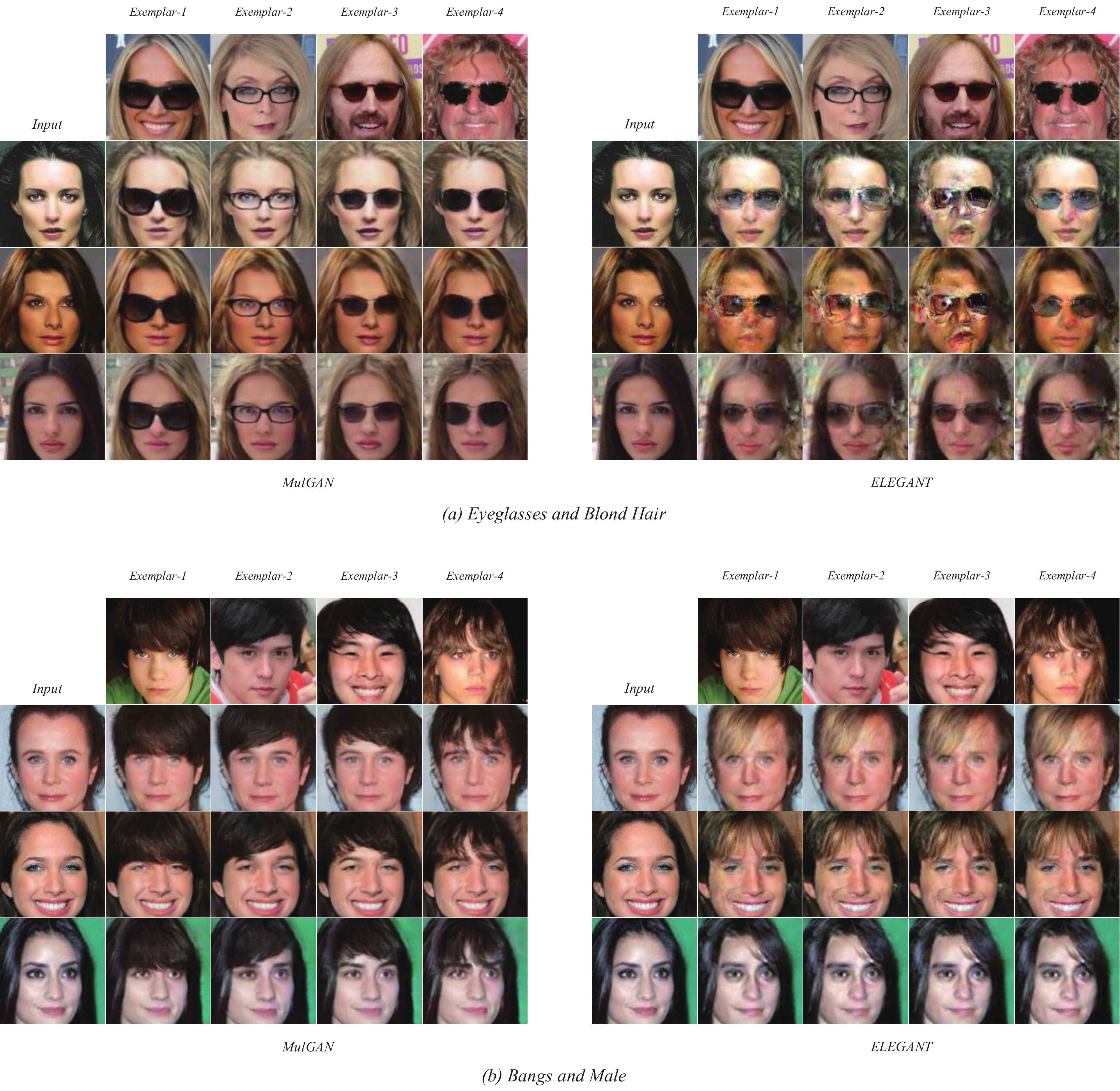}
	\caption{Comparisons of multiple facial attribute editing by exemplars. Panels from left to right are results of MulGAN and ELEGANT.\label{F5}}
\end{figure*}

\begin{figure}[htb]
	\centering
	\includegraphics[width=0.4\textwidth]{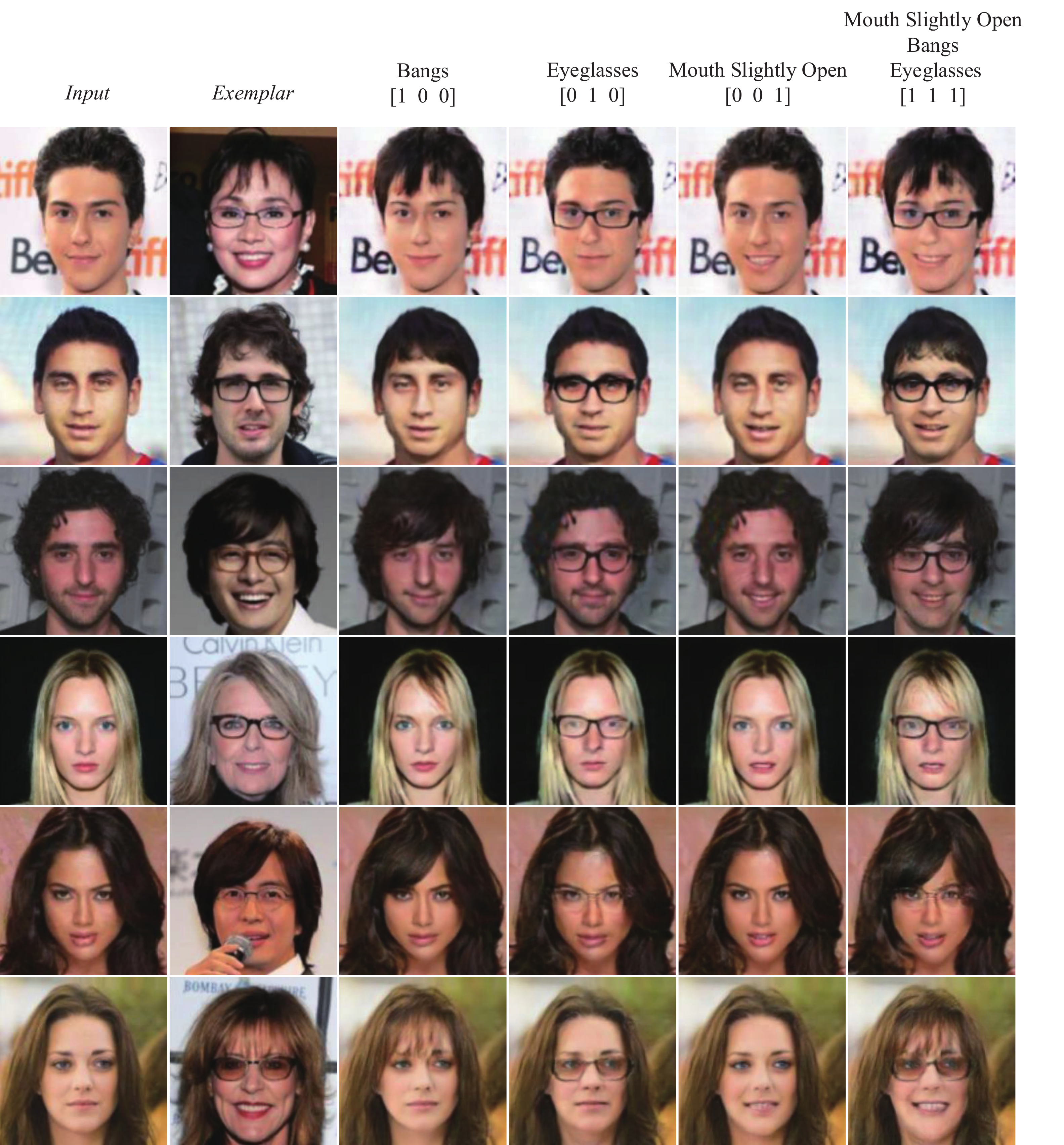}
	\caption{Illustration of controlling transferring attribute by labels. Zoom in for better resolution.\label{F6}}
\end{figure}
\begin{table}[htbp]
	\caption{Comparisons with DNA-GAN, ELEGANT and our MulGAN in terms of attribute style correlation.\label{Table1}}
	\centering
	\renewcommand\arraystretch{1.8}
	\begin{tabular}{cccc}
		\hline
		\tabincell{c} {\textbf{Attribute} \\  \textbf{style} \\ \textbf{correlation}}	& \textbf{Bangs}	& \textbf{Smiling}	& \textbf{Eyeglasses}\\
		\hline
		DNA-GAN    					& 0.95			& 0.9				  			  & 0.9 \\
		ELEGANT						& 0.6			& 0.7							  & 0.65 \\
		MulGAN  					& 0.95			& 0.85							  & 0.85 \\
		\hline
		
	\end{tabular}
\end{table}
\begin{table}[htbp]
	\caption{Comparisons with DNA-GAN, ELEGANT and our MulGAN in terms of FID.\label{Table2}}
	\centering
	\renewcommand\arraystretch{1.8}
	\begin{tabular}{cccc}
		\hline
		\textbf{FID}	& \textbf{Bangs}	& \textbf{Smiling}	& \textbf{Eyeglasses}\\
		\hline
		DNA-GAN    					& 82.51			& 79.43				  			  & 80.74 \\
		ELEGANT						& 35.46			& 31.88							  & 46.85 \\
		MulGAN  					& 24.33			& 23.12							  & 30.27 \\
		\hline
		
	\end{tabular}
\end{table}
\begin{figure}[htbp]
	\centering
	\includegraphics[width=0.5\textwidth]{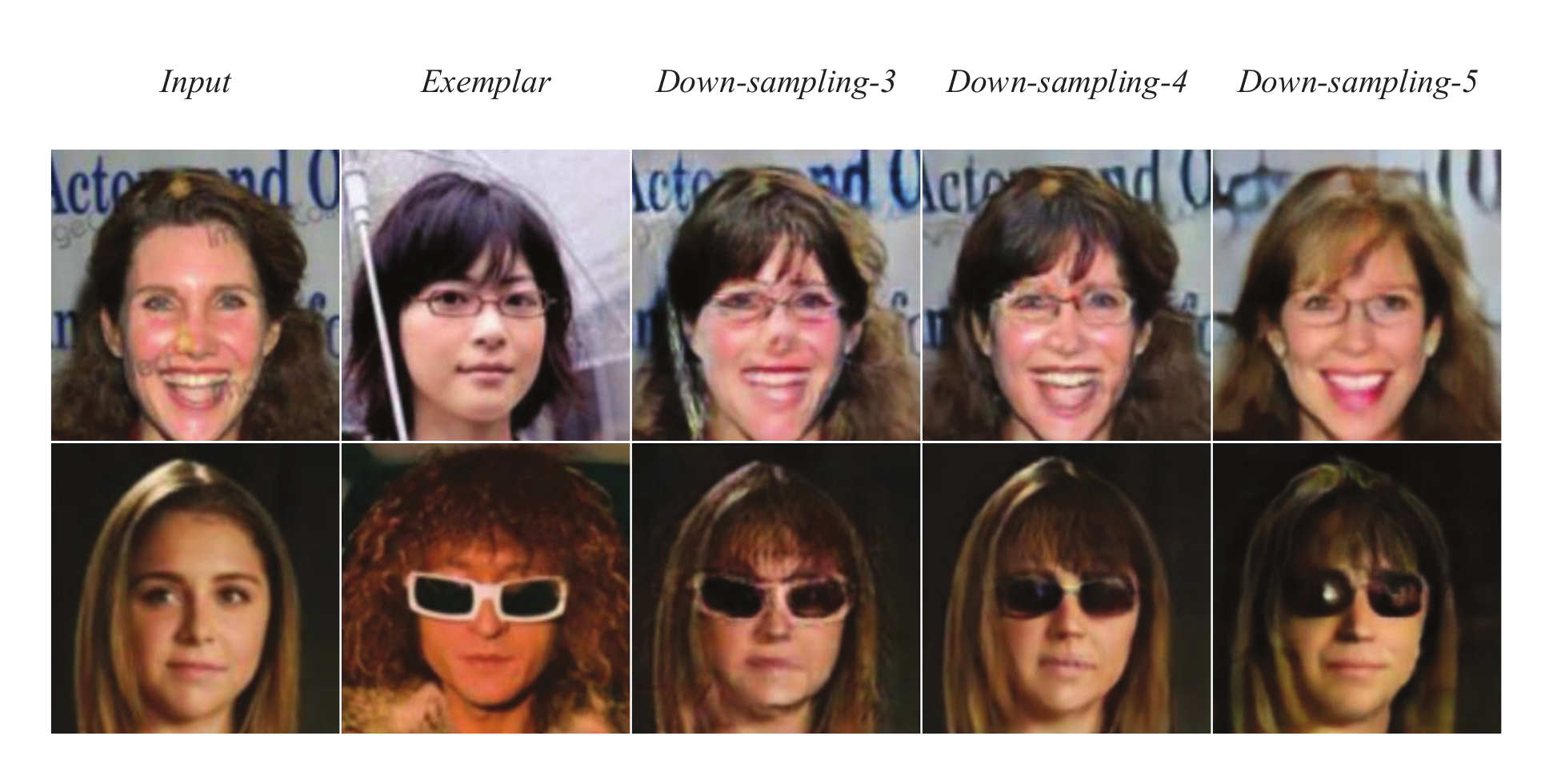}
	\caption{Comparison of training MulGAN with different down-sampling layer configurations. the facial attribute editing by exemplar here are Bangs and Eyeglasses.\label{F7}}
\end{figure}
\subsection{Dealing with Multiple Attributes Simultaneously}
We compare our methods with ELEGANT for transferring multiple attributes of exemplar simultaneously. For ELEGANT, we achieve transferring the multiple attributes of the exemplar by multiple exchange operations. For example, ELEGANT want to transfer “Pale Skin” and “Smiling” two attributes by exemplar. First ELEGANT transfer “Pale Skin” attribute by exchanging their attribute-related latent feature region of input image and exemplar, which can generate a new image with “Pale Skin” attribute, and then ELEGANT transfer “Smiling” attribute based on the new generated image by the same exemplar. By this way, ELEGANT can transfer two attributes from an exemplar to the input image. Note that the order of attribute transfer is arbitrary, which does not affect the generated results. Two models are performed on the same face images and reference images with respect to three attributes. As shown in Fig.~\ref{F5}, no matter the capability of transferring the attribute by exemplars or the quality of the generated image, our method is completely better than ELEGANT. (zooming in for a closer look).

\subsection{Quantitative evaluation}
Like other image generation tasks, facial image editing by exemplars lacks good quantitative evaluation metrics. Attribute generation accuracy is not a good metric for evaluating facial attribute editing by exemplar as it mostly focuses on classification accuracy on the changed attributes of the generated image, not on the style of attributes transferred from the exemplar into the generated image. So we perform a user study using the test set of the CelebA dataset for evaluating the relevance of transferred attribute style between the generated image and the exemplar image. The 20 users are shown both the exemplar image and the generated image by exemplar, and are asked to guess if the attribute style is relevant between the generated image and the exemplar. Table~\ref{Table1} shows the percentage of the generated images that are deemed to be relevant to the exemplar on the transferred attribute style. It can be seen that our MulGAN outperforms all the competing methods for the representative attributes Bangs, Smiling and Eyeglasses. In addition, in order to evaluate the quality of generated image, we employ FID~\cite{23} to quantify the distance from the distribution of synthesized images to the real ones, computed in the feature spaces of the retrained Inception Network~\cite{24}. Lower FID values indicate better quality of the generated images, as the two distributions are closer to each other. Table~\ref{Table2}. lists the FID values on three representative facial attributes. It is clear that our approach achieves competitive results compared with other methods.

\subsection{Controllable Attribute Transfer by Labels}
Directly applicable for controlling transferring attribute by labels is a characteristic of our MulGAN. For a given exemplar, the user can choose the single or multiple facial attributes of interest by binary attribute labels to transfer from the exemplar to the input image. As shown in Fig.~\ref{F6}, the exemplars own three attribute: Bangs, Eyeglasses and Mouth Slightly Open. The user only need set the value to 1 in the corresponding position of the binary attribute label vector if wanted to transfer the attribute of interest, otherwise set to 0. Then MulGAN filters the attribute-relevant blocks based on a user's given binary labels and generates a new image with the same style of attribute as the exemplar.

\subsection{Ablation Study}
We train our MulGAN with three, four, and five down-sampling layers, respectively to investigate the influence of down-sampling layers for face attribute editing by an exemplar. The results are shown in Fig.~\ref{F7}. As we can see, lots of details in the exemplar are transferred to the input image when three down-samplings layers is used, which includes not only attribute-related details, but also attribute-irrelevant details. The attribute-related details make the generated image own exactly the same style of attribute as the exemplar,  which can enhance the attribute transferring ability of the model by exemplars. However, the attribute-irrelevant details seriously interfere with the quality of the generated image as shown in Fig.~\ref{F7}. Besides, fewer down-sampling layers lead to model with poor generation ability, as a result, the model cannot fuse the transferred attribute-related details with the original image attribute-irrelevant information well. When five down-samplings layers is used, MulGAN can obtain high-level abstraction of image content and attribute, but lots of details are lost. Loss of the attribute-irrelevant details in the input image seriously affects the quality of the generated images and gives rise to blurry and low quality result as shown in Fig.~\ref{F7}. Loss of the attribute-relevant details in the exemplar causes failure to accurately transfer the same style of attribute in the exemplar, the exemplar can only act as labels in extreme cases. But More down-samplings layers make model own stronger generation ability, which can fuse the transferred attribute-related details with the input image attribute-irrelevant information well. When four down-samplings layers is used. MulGAN can achieve a better balance between the ability of transferring attributes by exemplar and the quality of generated images.

\section{Conclusion}
In this paper, we proposed a novel model MulGAN for transferring multiple face attributes by an exemplar, which can extract the attribute information of image into the predefined different blocks by introducing the binary attribute labels and exerting the attribute classification constraint. Under the observation that excessive down-sampling layers causes the loss of lots of details, resulting in failing to transfer the style of attribute in the exemplar and generating blurry and low quality results. we adopt the less down-sampling layers to enhance the attribute transfer ability and improve the image quality. Experiments demonstrate that MulGAN can accurately transfer the style of multiple face attributes in the exemplar, while well preserving the attribute-excluding details, with better visual effect than the competing methods.

\section*{Acknowledgments}
The authors acknowledge support from the Natural Science Foundation of China (No.61300072, 31771475).

\bibliographystyle{ieeetr}

\end{document}